\documentclass{article}

\usepackage{amsmath}
\usepackage{comment}
\usepackage{booktabs}
\usepackage{booktabs}
\usepackage{graphicx}

\usepackage{wrapfig}

\usepackage{xcolor}

\newcommand{\luca}[1]{\textcolor{black}{#1}}
\newcommand{\circled}[1]{\textcircled{\scriptsize #1}}

\usepackage{booktabs}
\usepackage{multirow}
\usepackage{graphicx}
\usepackage{subcaption}

\renewcommand{\thefootnote}{\fnsymbol{footnote}}

\usepackage[final]{corl_2026} 

\title{Before Parc Fermé: RL-Time Pruning for\\Efficient Embodied LLMs in Autonomous Driving} 

\author{
  \textbf{Luca Benfenati}$^{1,*}$,
  \textbf{Ali Azimi}$^{1,*}$,
  \textbf{Matteo Risso}$^{1,3}$,\\
  \textbf{Fabio Carapellese}$^{2}$,
  \textbf{Daniele Jahier Pagliari}$^{1}$,
  \textbf{Alessio Burrello}$^{1}$\\[0.5em]
  $^{1}$Department of Control and Computer Engineering, Politecnico di Torino\\
  $^{2}$Department of Mechanical and Aerospace Engineering, Politecnico di Torino\\
  $^{3}$EIS Lab, The Italian Institute of Artificial Intelligence (AI4I)\\  
}

\begin{document}
\maketitle
\footnotetext[1]{Denotes equal contribution.}
\setcounter{footnote}{0} 
\renewcommand{\thefootnote}{\arabic{footnote}}


\begin{abstract}
Embodied Large Language Models (LLMs) are increasingly used as reasoning modules in robotic control pipelines to improve human-robot interaction, but their memory and generation latency make real-time deployment difficult. Pruning can reduce these costs, but for controllers that undergo multiple pre- and post-training phases, the crucial question is not only how much to prune, but when pruning should occur.

In this work, we propose \emph{Before Parc Fermé} (BPF), a pruning strategy performed during RL that compresses embodied LLM controllers while they are still being optimized for closed-loop behavior. This allows pruning decisions to account for the task-specific supervision and closed-loop feedback that shape the final controller. We propose two variants: BPF-RL, which performs iterative pruning during RL by removing part of the model at predefined training intervals, and BPF-SFT/RL, which first prunes part of the model structure during SFT and then further compresses it during RL using the same iterative strategy as BPF-RL until the target pruning ratio is reached.
We evaluate BPF on RobotxR1, an LLM-based autonomous-driving control pipeline, using an established LLM pruning framework (LLM-Pruner), and compare it against post-training pruning, post-training pruning with RL recovery, SFT-stage pruning, and smaller dense models from the same family.

Our results show that BPF provides the best task-performance vs. memory and throughput trade-off among the considered pruning strategies. When compressing the larger RobotxR1 models, BPF-SFT/RL achieves a $1.69\times$ better size--end-to-end performance trade-off than directly selecting a smaller dense model from the same family, measured as removed parameters per lost percentage point of control adaptability. On the Jetson AGX Orin mounted on the target robotic platform, the compact models improve decode throughput by up to $27\%$. Finally, we further show that pruning must be evaluated end-to-end, since changes in generated output tokens can affect downstream latency and closed-loop behavior.
\end{abstract}


\section{Introduction}
Large language models (LLMs) are increasingly used in robotic systems as high-level interfaces between human commands and machine actions~\cite{palm_saycan, code_as_policies, inner_monologue, llm_planner, ismail2024narrate}.
Through large-scale pretraining, these models acquire broad linguistic, task-level, and commonsense knowledge, which makes them useful for interpreting underspecified natural-language commands and adapting them to the current robotic context.
However, even relatively small LLMs introduce non-negligible memory, latency, and energy overheads, making them a difficult target for real-time local robotic deployment~\cite{eff_vla_survey}.

Recently, RobotxR1~\cite{robotxr1} studied the use of LLMs for autonomous driving control, where the model is used to translate natural language user-specified driving instructions into closed-loop vehicle behavior. RobotxR1 uses two LLM-based modules: DecisionxR1 evaluates whether the current driving behavior matches the user instruction, while MPCxR1 adapts Model Predictive Control (MPC) parameters to realize the desired behavior. 
Both modules are trained with supervised fine-tuning (SFT) followed by GRPO-based~\cite{deepseek-r1} reinforcement learning (RL), allowing their behavior to improve through interaction with the driving environment.

Despite the competitive task performance of RobotxR1~\cite{robotxr1}, the gap highlighted in the results between larger and smaller models like Qwen2.5-3B and Qwen2.5-1.5B~\cite{hui2024qwen2} illustrates a central deployment problem: the larger model improves control adaptability by more than 20\%, but increases LLM token generation throughput by more than 60\%, making real-time deployment harder. This creates a trade-off between the model scale that learns robust embodied reasoning and the one that can be efficiently executed on a constrained robotic platform. 
Pruning is a natural way to address this issue, since it can reduce the memory and computational cost of larger LLMs while retaining part of their task-performance advantage over smaller manually-designed models.

However, the correct way to apply it in an embodied LLM system trained with SFT and RL is not obvious and has not yet been studied extensively. 
If pruning is applied after SFT but before RL, the pruning criterion has no access to the closed-loop feedback that determines which structures matter for the final driving task.
If pruning is applied only after SFT and RL, compression acts on the final task-adapted model, but the remaining weights have limited opportunity to adapt to the removed structures. 
Pruning during RL addresses both issues: the model has already learned the supervised task format, and further closed-loop feedback is available before training has fully converged.

Based on this observation, we propose \emph{Before Parc Fermé}\footnote{\emph{Parc fermé} refers to the Formula 1 period in which cars are locked from further modification before the race. Here, it denotes the post-training stage where the model architecture is fixed.} (BPF), 
a family of RL-coupled pruning recipes for embodied LLM controllers.
BPF compresses the model while it is still being optimized for closed-loop behavior, rather than treating pruning as a purely post-training operation.
We study two variants: \emph{BPF-RL}, which performs iterative pruning during RL, and \emph{BPF-SFT/RL}, which first prunes part of the model structure during SFT and then further compresses the remaining structures during RL using the same iterative strategy as BPF-RL until the target compression ratio is reached.
Our central research thesis is that a larger LLM compressed with BPF during adaptation can preserve more reasoning and control ability than models pruned at other points of the SFT-RL pipeline, including before RL and after training.

We evaluate BPF on both RobotxR1 LLM-based modules, DecisionxR1 and MPCxR1. 

Our contributions are summarized as follows. 
\textbf{(i)} We introduce BPF, a pruning recipe that couples RL and structured pruning for embodied LLM controllers, using RobotxR1 as a representative SFT/RL-trained robotic LLM system. We show that this strategy is more effective than pruning only after training, pruning before RL, or applying post-RL recovery. Our best BPF-SFT/RL strategy improves task performance by up to 11.4\% with respect to doing pruning and post-RL recovery.
\textbf{(ii)} We characterize the resulting deployment trade-off in terms of task performance, memory footprint, generation throughput, and latency.
In particular, BPF-SFT/RL achieves a $1.69\times$ better size--end-to-end performance trade-off than directly selecting the smaller Qwen2.5-1.5B model, measured as removed parameters per lost percentage point of control adaptability, while improving decode throughput by $27\%$ on the Jetson AGX Orin mounted on the target robotic platform.
\textbf{(iii)} We show that pruning embodied LLM controllers must be evaluated end-to-end, not only through task-agnostic metrics (e.g., memory footprint, generation throughput). Since LLM latency depends on the generated-token count, which can change after pruning, compression can alter downstream latency and behavior. This is especially relevant in RobotxR1, where DecisionxR1 outputs condition MPCxR1 generation.
\section{Related Work}\label{sec:related}

\paragraph{LLMs for Robot Control and Autonomous Driving:}
Recent work has studied LLMs as high-level reasoning modules for robotic systems, where the language model interprets instructions, decomposes tasks, selects skills, or generates objectives for a downstream controller \cite{palm_saycan, code_as_policies, inner_monologue, llm_planner, ismail2024narrate}. 
In autonomous driving, LLMs and multimodal LLMs have been used for scene interpretation, decision-making, and planning \cite{drivegpt4, drivelm, drive_as_you_speak}. 
\citet{onboard_llm_ads} introduces the on-board LLM-based autonomous-driving pipeline used in this work, where a decision model interprets driving instructions, and an MPC-oriented language model adapts controller parameters. 
\citet{robotxr1} extend this pipeline by adapting the \textit{R1-Zero} training protocol~\cite{deepseek-r1} to the same closed-loop autonomous-driving environment, bringing reasoning-oriented LLMs into the robotic control setting and showing that they outperform their non-reasoning counterparts.
Building on this setup, our work studies whether pruning during learning is a better alternative to replacing the model with a smaller dense one from the same family, and investigates the optimal training stage for pruning.

\paragraph{Post-Training LLM Pruning:}
Recent LLM-specific pruning methods reduce model size and inference cost by applying unstructured or structured sparsification to pretrained models \cite{sparsegpt, wanda, llm_pruner, flap, slicegpt, RL-pruner}. 
In particular, RL-Pruner~\cite{RL-pruner} further uses RL to recover from pruning-induced loss in encoder-based language models without retraining.
While all these methods provide practical mechanisms for compressing LLMs, they treat pruning as a compression step applied to an already-trained model, rather than as part of task adaptation or reasoning-oriented training.

\paragraph{Learning-Aware LLM Pruning:}
A more relevant direction for our work is pruning that is integrated with training or fine-tuning. 
Sheared LLaMA~\cite{sheared_llama} uses targeted structured pruning to reduce a larger pretrained LLM to a specified smaller architecture, followed by continued pretraining, showing that starting from a larger model can be more effective than training a smaller model directly. 
Pruning-Aware Tuning~\cite{pat} incorporates structured pruning into finetuning by learning pruning decisions jointly with task adaptation, rather than fixing the sparsity pattern before finetuning.
Adapt-Pruner~\cite{adapt_pruner} further studies adaptive structured  pruning during training for efficient small language model construction. 
Closest in spirit to our setting, reinforcement learning has also been used to learn structured pruning distributions for CNN compression~\cite{rl_pruner_cnn}, however, this work optimizes a pruning policy for CNNs, rather than pruning an LLM during RL-based embodied control training.

Our work studies learning-aware pruning in embodied robotics~\cite{robotxr1} and in an \textit{R1-Zero}-style setup~\cite{deepseek-r1}, where the model is adapted through SFT and closed-loop RL to acquire reasoning behavior. 
We study pruning as a training-time design choice, comparing compression after training, during SFT, during RL, and across both stages.
This distinction matters because the relevant costs in robotics are not only model size but also inference latency, generated token length, and downstream effects across the control pipeline.
\section{Methodology}
\label{sec:method}
\begin{figure}[ht]
    \centering
    \includegraphics[width=\linewidth]{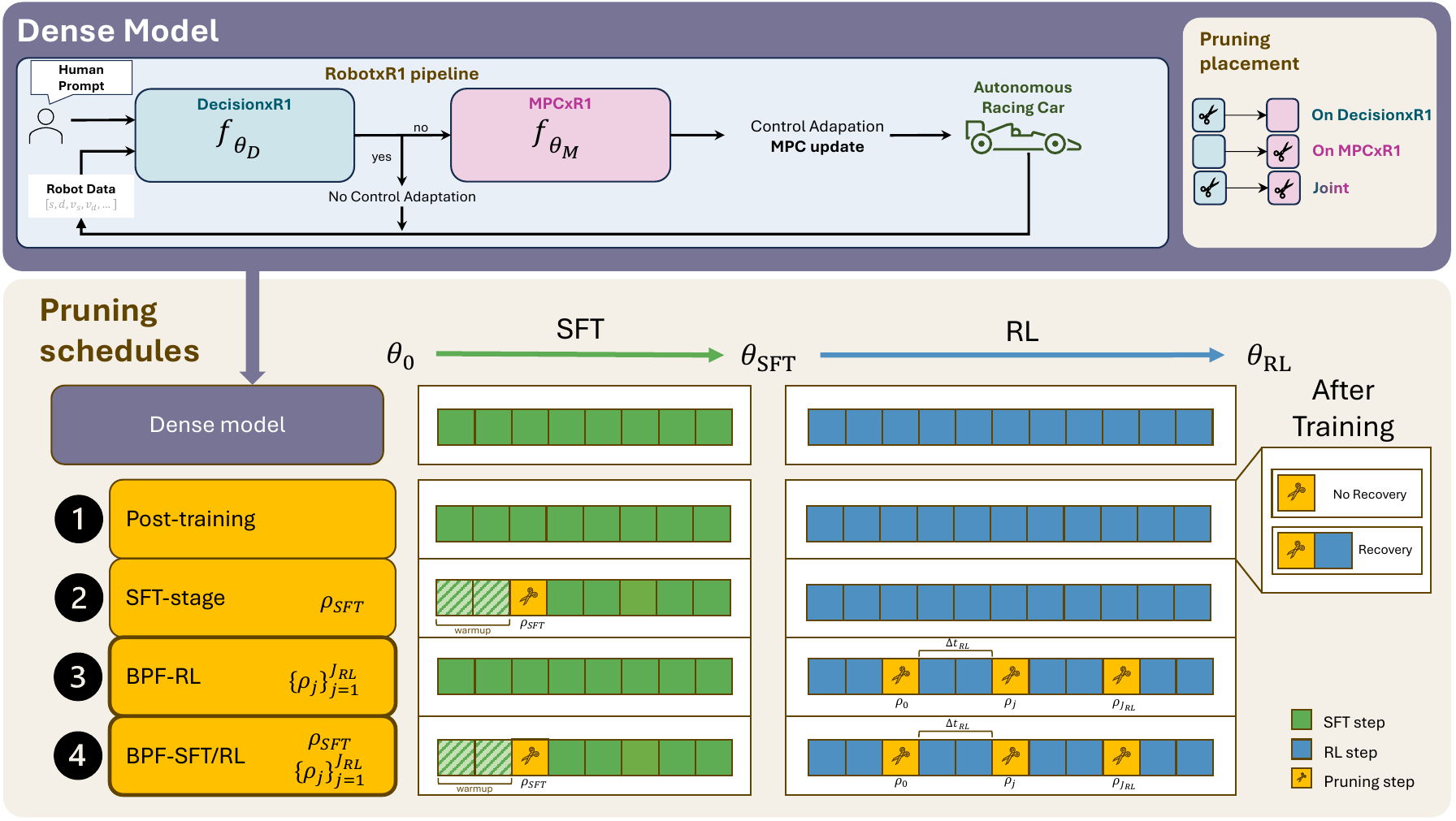}
    \caption{
    Overview of the pruning strategies and placement in our pipeline.
    }
    \label{fig:method_overview}
\end{figure}
This work builds on RobotxR1~\cite{robotxr1}, which adapts LLMs for robotic reasoning through SFT followed by closed-loop RL.
We keep the controller, prompts, rewards, and evaluation protocol unchanged, and study how structured pruning can be integrated into the same training pipeline.
The key methodological question is where to apply pruning so that compression can exploit closed-loop training feedback without being delayed until adaptation has fully converged.

\subsection{RobotxR1 Pipeline}
\label{subsec:decision_mpc_pipeline}
We use the same DecisionxR1-MPCxR1 pipeline as RobotxR1~\cite{robotxr1}.
DecisionxR1 is a RAG-enhanced LLM that receives the user command and recent robot state history and determines whether the current vehicle behavior matches the requested behavior.
The state history contains track-relative information such as vehicle position with respect to the racing line, heading error, steering angle, and longitudinal velocity.
MPCxR1 then uses the DecisionxR1 output, the user command, the robot state history, retrieved context, and the MPC formulation to generate textual updates to the exposed MPC parameters.
The parsed parameters are applied to the low-level MPC controller, which executes the corresponding behavior in closed loop.

Both modules follow the same high-level adaptation sequence with SFT followed by RL.
During SFT, the models are trained with a next-token prediction loss on task-specific instruction-response data.
During RL, DecisionxR1 is optimized with behavior-adherence and formatting rewards, while MPCxR1 is optimized through closed-loop simulation using driving-performance, formatting, and parameter-validity rewards.
This shared SFT-RL structure defines the training pipeline in which we introduce pruning.
Additional details on the inherited RobotxR1 module inputs, outputs, and rewards are reported in Appendix~\ref{app:robotxr1_details}.

Because DecisionxR1 and MPCxR1 are executed sequentially, pruning can have both local and downstream effects.
Pruning DecisionxR1 can change the text received by MPCxR1, while pruning MPCxR1 directly affects controller-parameter generation.
In the experiments, we therefore apply the same pruning schedules to DecisionxR1, MPCxR1, and both modules jointly.
\subsection{BPF Pruning Schedule}
\label{subsec:bpf_pruning}
BPF defines when pruning is introduced within the SFT-RL training pipeline.
Since DecisionxR1 and MPCxR1 share this training structure, we define all schedules for a generic LLM module with weights $\theta$ and apply them to either module in the experiments.

Following Fig.~\ref{fig:method_overview}, we compare pruning after the full SFT-RL pipeline, during SFT, during RL, and across both stages.
Our main focus is on the two RL-coupled schedules: incremental pruning during RL (BPF-RL), and split SFT-RL pruning (BPF-SFT/RL), where part of the model is pruned during SFT, and the remaining compression follows the same incremental RL schedule as BPF-RL.

\textbf{Post-training pruning.}
The simplest pruning schedule is \textbf{post-training} pruning (PTP), denoted as \circled{1} in Fig.~\ref{fig:method_overview}, where the model is first adapted with the standard SFT and RL, obtaining $\theta_{\mathrm{RL}}$, and is then pruned in one-shot. This corresponds to the conventional setting in which compression is applied only after task adaptation has completed. We consider both the direct post-training variant (PTP) and a recovery variant (PTP+R), where the pruned model is further optimized with additional RL steps to recover the lost performance.

\textbf{SFT-stage pruning.}
Fig.~\ref{fig:method_overview}-\circled{2} depicts \textbf{SFT-stage} pruning schedule, where pruning is applied as a single compression step during SFT, before the RL stage starts. Starting from $\theta_0$, the model is first trained for an SFT warmup phase without pruning. After this warmup, the model is pruned with target ratio $\rho_{\mathrm{SFT}}$, and the remaining SFT steps continue on the compressed model, yielding $\theta_{\mathrm{SFT}}^{\rho}$. RL is then run on this already compressed model.

\textbf{BPF-RL.}
Fig.~\ref{fig:method_overview}-\circled{3} depicts our first RL-coupled pruning schedule, \textbf{BPF-RL}.
The model is first trained with SFT without pruning, obtaining $\theta_{\mathrm{SFT}}$.
Compression is then introduced during RL, so that pruning decisions are made after the model has learned the supervised task format, but while it is still being optimized with closed-loop feedback.
Rather than pruning once, BPF-RL removes weights iteratively during RL.
Let $\rho$ be the final target pruning ratio and let $J_{\mathrm{RL}}$ be the number of pruning events.
We split the target compression uniformly across pruning events, so that each event increases the cumulative pruning ratio by $\rho/J_{\mathrm{RL}}$.
Starting from $\theta_{\mathrm{SFT}}$, RL is run for an initial warm-up interval.
Then, every $\Delta t_{\mathrm{RL}}$ RL steps, pruning scores are recomputed on the current model, and the lowest-scoring parameter groups are removed until the cumulative pruning ratio reaches $j\rho/J_{\mathrm{RL}}$ at pruning event $j$.
After each pruning event, RL resumes under the same reward objective.
This alternation between RL updates and pruning allows the remaining weights to adapt to the removed structures while the model is still being optimized for the driving task.

\textbf{BPF-SFT/RL.}
Fig.~\ref{fig:method_overview}-\circled{4} shows our second RL-coupled pruning schedule, \textbf{BPF-SFT/RL}.
This schedule is motivated by the fact that SFT and RL provide complementary information.
SFT exposes the model to task-specific prompts, output formats, and supervised input-output behavior, making it a useful stage to remove clearly redundant structure.
However, SFT does not provide closed-loop feedback from the final driving task.
For this reason, BPF-SFT/RL applies only a partial compression during SFT and leaves the remaining compression to RL, where pruning scores can reflect the model after it has started adapting to closed-loop behavior.

Concretely, let $\rho$ be the final target pruning ratio.
BPF-SFT/RL first prunes the model during SFT up to an intermediate ratio $\rho_{\mathrm{SFT}} < \rho$, after an SFT warm-up phase.
The remaining compression, $\rho - \rho_{\mathrm{SFT}}$, is then reached during RL using the same iterative procedure as BPF-RL.
If $J_{\mathrm{RL}}$ pruning events are used during RL, each event adds $(\rho-\rho_{\mathrm{SFT}})/J_{\mathrm{RL}}$ to the cumulative pruning ratio.
Thus, BPF-SFT/RL combines partial compression after supervised task adaptation with RL-time compression, allowing the model to access task feedback that determines the final control behavior.

\luca{Appendix~\ref{app:iter_sft} additionally considers iterative pruning performed solely during SFT instead of during RL as in BPF-RL and BPF-SFT/RL, separating its effect from where it is placed. This additional approach is proven to be suboptimal when compared to ours BPF pruning schedules.}
\subsection{Taylor-Based Pruning}
\label{subsec:taylor_pruning}
The pruning schedules described in previous sections are agnostic to the specific pruning criterion and can be instantiated with either structured or unstructured pruning.
In this work, to demonstrate BPF performance, we adopt LLM-Pruner~\cite{llm_pruner}, an LLM pruning framework that can operate at different granularities. In the structured setting, the method first identifies groups of dependent parameters that must be removed jointly to preserve architectural consistency after compression. In the unstructured setting, the same formulation is recovered by treating each scalar weight as an independent group.

Given a calibration batch of data $\mathcal{D}$ and a loss $\mathcal{L}(\mathcal{D};\theta)$ evaluated at the current parameters $\theta$, the importance of a scalar parameter $w_i \in \theta$ is approximated with a first-order Taylor score:
\begin{equation}
    I(w_i)
    =
    \left|
    w_i
    \frac{\partial \mathcal{L}(\mathcal{D};\theta)}{\partial w_i}
    \right|
    \label{eq:taylor_score}
\end{equation}
For a coupled group $\mathcal{G}_k$, we aggregate the scores of its parameters as $S(\mathcal{G}_k) = \sum_{w_i \in \mathcal{G}_k} I(w_i)$.
Groups with the lowest scores are removed until the target pruning ratio for that pruning point is reached.

In this work, we focus on structured pruning because it can more straight-forwardly yield practical inference speedups~\cite{cfsp,slicegpt}. This better aligns with common deployment settings for standard inference runtimes and hardware, whereas unstructured sparsity often requires dedicated kernels and hardware to achieve practical speedups. We also report unstructured pruning results in Appendix~\ref{app:unstructured}, while \luca{Appendix~\ref{app:wanda} evaluates Wanda~\cite{wanda} as an alternative pruning criterion.}
\section{Experiments and Results}
\label{sec:Results}
%
All experiments are conducted on the RobotxR1 pipeline~\cite{robotxr1}, using the same controller, prompts, rewards, and closed-loop evaluation protocol described in Sec.~\ref{subsec:decision_mpc_pipeline}. Additional setup details are provided in Appendix~\ref{app:experimental_setup}.

\subsection{DecisionxR1 Pruning with Fixed MPCxR1}~\label{subsec:decision_pruning_fixed_mpc}
We first isolate the effect of pruning DecisionxR1, while leaving MPCxR1 unpruned. Following the control-adaptability metric used in RobotxR1, we report the average percentage improvement over the default MPC controller across four task errors, $(E_c,E_v,E_r,E_s)$. Here, the default MPC is the original hand-tuned racing-line controller before any LLM-generated parameter update. Additional details are provided in Appendix~\ref{app:robotxr1_details}.

%
\begin{wrapfigure}{r}{0.6\textwidth}
    \centering
    \vspace{-.8cm}
    \includegraphics[width=\linewidth]{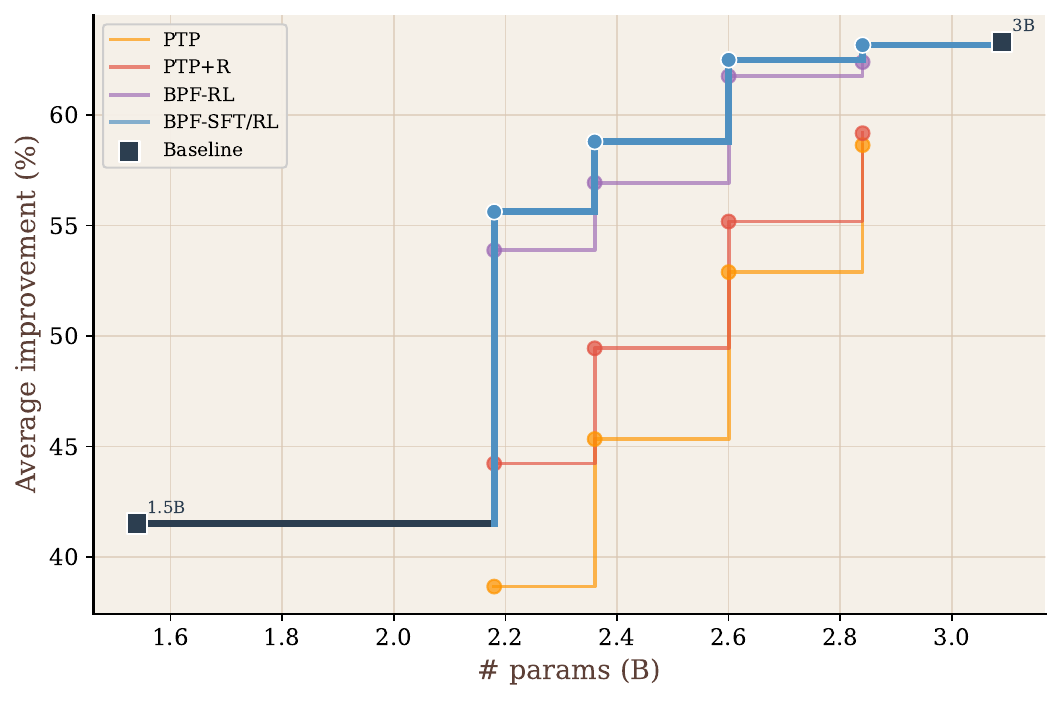}
    \vspace{-.5cm}
    \caption{
    Average control-adaptability improvement over the default MPC controller as in~\cite{robotxr1} versus model size.
    \vspace{-.5cm}
}
    \label{fig:decision_pareto}
\end{wrapfigure}

Fig.~\ref{fig:decision_pareto} shows the resulting trade-off between the DecisionxR1 model size and end-to-end control adaptability. The $x$-axis reports the number of parameters, while the $y$-axis reports the average improvement over the default MPC controller. Square markers denote the unpruned baselines. 
Starting from Qwen2.5-3B, we test $10\%$, $20\%$, $30\%$, and $45\%$ pruning ratios for PTP (orange), PTP+R (red), BPF-RL (purple), and BPF-SFT/RL (blue). SFT-stage pruning is omitted because it fails the task even at the smallest pruning ratio. Additional results obtained with best BPF-SFT/RL recipe starting from 1.5B baseline are reported in Appendix~\ref{app:structured}.

The results show that our novel BPF-RL and -SFT/RL give the best trade-off among the schedules applied to the 3B DecisionxR1 model. In particular, for the smallest architecture, obtained at $45\%$ pruning ratio, BPF-RL improves the average control-adaptability score by 9.7\%/15.2\% over PTP+R/PTP, while BPF-SFT/RL improves it by 11.4\%/17\%.
The same point also provides the best trade-off between the dense 3B and 1.5B models, achieving $0.91$B  parameter reduction with 7.61\% control-adaptability loss relative to the dense 3B model, corresponding to a ratio of $0.12$B parameter reduction per lost percentage point. By contrast, moving directly to the dense 1.5B backbone gives $1.55$B parameter reduction losing 21.7\% points, yielding only $0.071$B parameter reduction per lost point. Thus, BPF-SFT/RL provides a $1.69\times$ better compression--adaptability trade-off than selecting the smaller baseline model.
Moreover, the BPF-SFT/RL model with a 10\% pruning ratio achieves the same level of control adaptability as the strongest RobotxR1 baseline while reducing its parameter count by 0.25B.
\subsection{Joint DecisionxR1-MPCxR1 Pruning}
\label{subsec:dec_mpc_pruning_res}

We next evaluate full-pipeline pruning, where both DecisionxR1 and MPCxR1 are pruned. Since DecisionxR1 output is passed to MPCxR1, this setting captures how compression propagates through the full RobotxR1 control pipeline.

\begin{wrapfigure}{r}{0.6\textwidth}
    \centering
    \includegraphics[width=\linewidth]{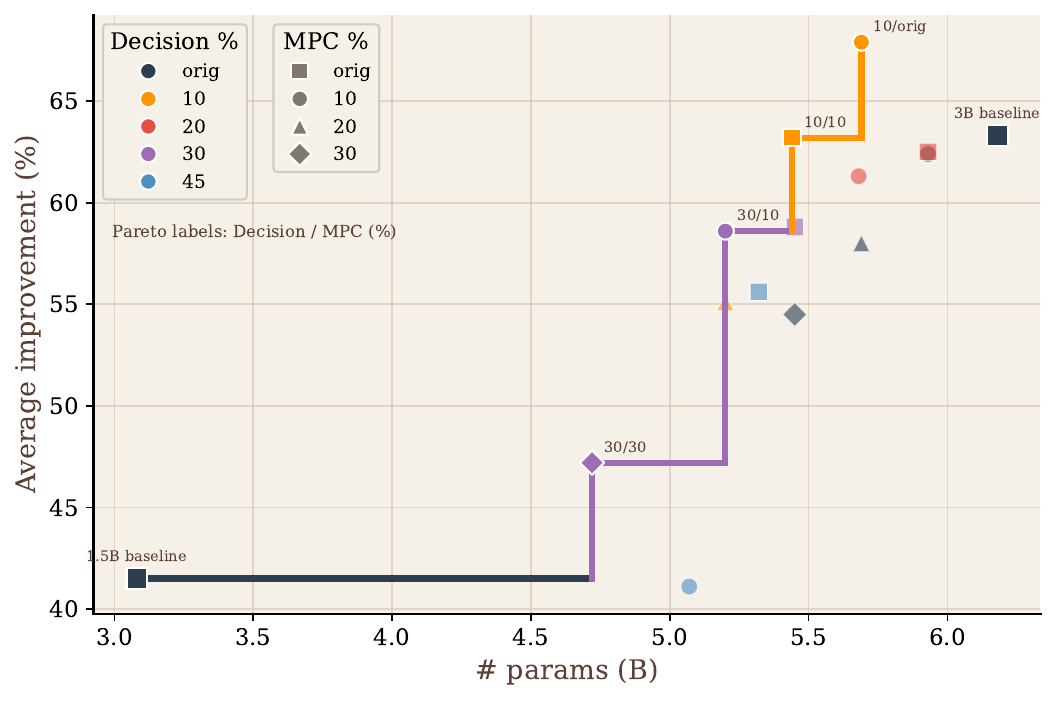}
    \caption{
    End-to-end control-adaptability trade-off when pruning both DecisionxR1 (colors) and MPCxR1 (shapes).
    }
    \label{fig:decision_mpc_pareto}
\end{wrapfigure}

Fig.~\ref{fig:decision_mpc_pareto} reports on the $x$-axis the total number of parameters across the two LLM modules, while the $y$-axis reports the average improvement over the default MPC controller. Different colors indicate the DecisionxR1 pruning ratio, while different marker shapes indicate the MPCxR1 pruning ratio with Pareto labels reporting DecisionxR1/MPCxR1 pruning ratios.
The labeled Pareto points show that moderate MPCxR1 pruning is critical. The $30/10$ configuration keeps the pipeline close to the unpruned 3B baseline, with less than $5\%$ loss in average improvement, whereas the $30/30$ configuration drops to roughly the mid-$40\%$ range. Thus, increasing MPCxR1 pruning from $10\%$ to $30\%$ causes a substantially larger degradation than increasing DecisionxR1 pruning alone. This confirms that MPCxR1 is the more sensitive module, consistent with its role in directly generating the MPC-parameter update used by the low-level controller.

The same trend holds when both modules are pruned jointly. Light compression of the full cascade can even improve over the unpruned 3B baseline. Indeed, the $10/10$ configuration achieves the best Pareto point, suggesting that BPF removes parameters that are not beneficial for the final closed-loop task. However, the gain is module-dependent. The best trade-off configurations keep MPCxR1 only mildly pruned, $10\%$, while allowing larger compression in DecisionxR1, up to $30$--$45\%$. 

\subsection{Embedded Deployment}\label{subsec:deployment}

\begin{wrapfigure}{r}{0.34\textwidth}
    \centering
    \vspace{-3.0em}
    \includegraphics[width=\linewidth]{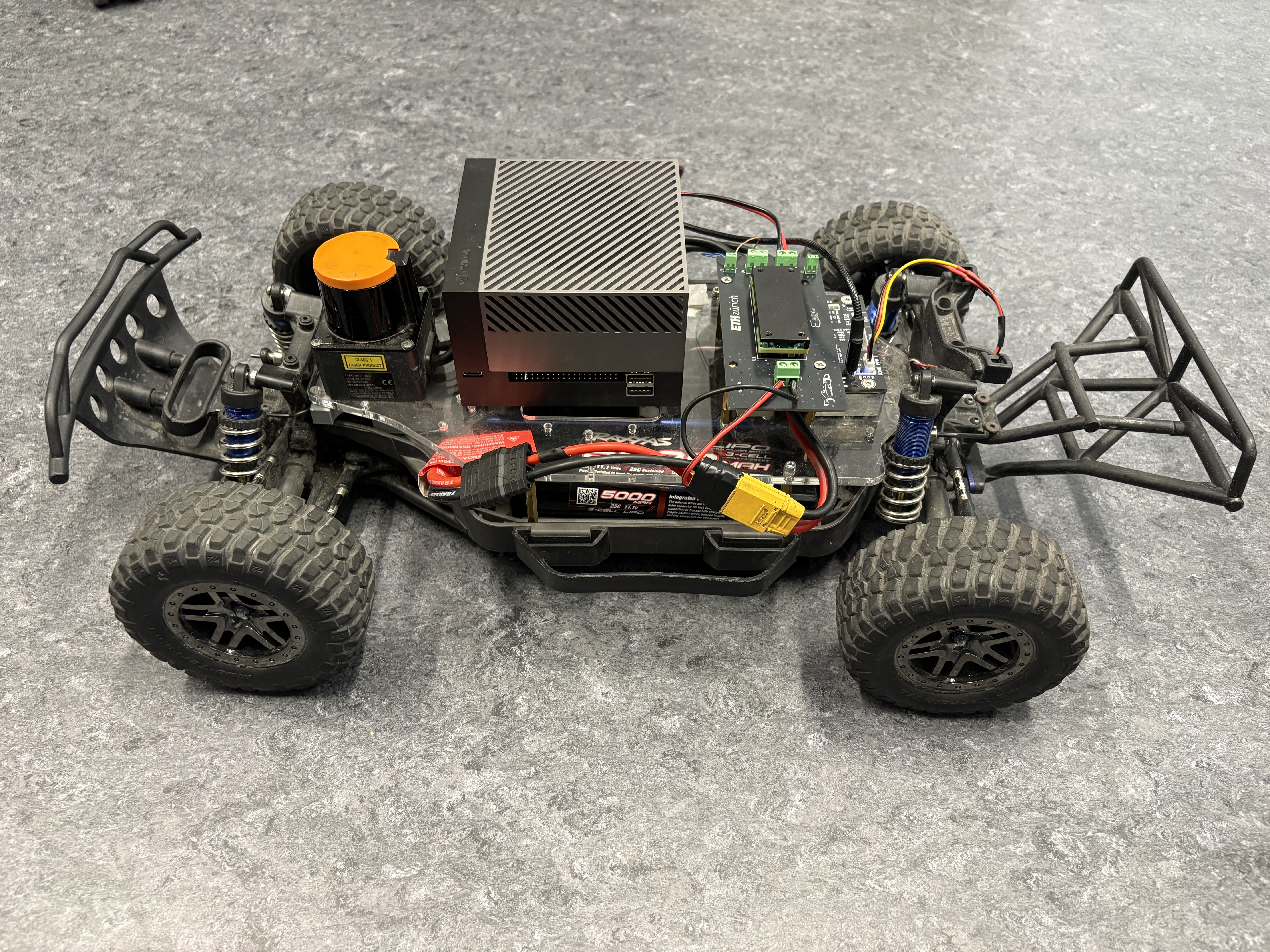}
    \caption{Robotic platform equipped with the Jetson AGX Orin used for deployment.}
    \label{fig:car_platform}
    \vspace{-1.0em}
\end{wrapfigure}
To assess whether our BPF models provide practical deployment benefits, we deploy all pruned configurations identified in Sec .~\ref {subsec:dec_mpc_pruning_res} on the physical robotic platform shown in Fig.~\ref{fig:car_platform}. The vehicle is equipped with a Jetson AGX Orin, which acts as the onboard inference device for the LLM modules. Additional measurements reported in Appendix~\ref{app:deployment}.

The deployment results are summarized in Table~\ref{tab:deployment}.
For fairness, all dense baselines are redeployed and remeasured on our deployment stack, where they achieve higher throughput than in the original RobotxR1 implementation~\cite{robotxr1}.
All pruned models are obtained with the BPF-SFT/RL recipe. After pruning, each model is quantized to \texttt{Q5\_K\_M}, exported to \texttt{GGUF} format, and deployed with the \texttt{llama.cpp} runtime~\cite{llamacpp}. Before deployment, we apply a compilation pass that zero-pads MLP weights so that output dimensions are multiples of $256$, avoiding FP16 fallbacks in \texttt{llama.cpp} for non-aligned layers. Memory footprint and decode throughput are measured with the \texttt{llama-bench} utility.

Table~\ref{tab:deployment} shows that BPF consistently improves inference throughput while reducing memory footprint. On the Jetson AGX Orin, pruning DecisionxR1 by $45\%$ reduces memory from $2.07$GB to $1.51$GB and increases decode throughput from $45$ to $58$ tok/s. A similar trend is observed for MPCxR1 pruning: at $30\%$ ratio, memory decreases to $1.62$GB and throughput reaches $55$ tok/s.

\begin{wrapfigure}{r}{0.6\textwidth}
    \centering 
    \vspace{-5em}
    \includegraphics[width=\linewidth]{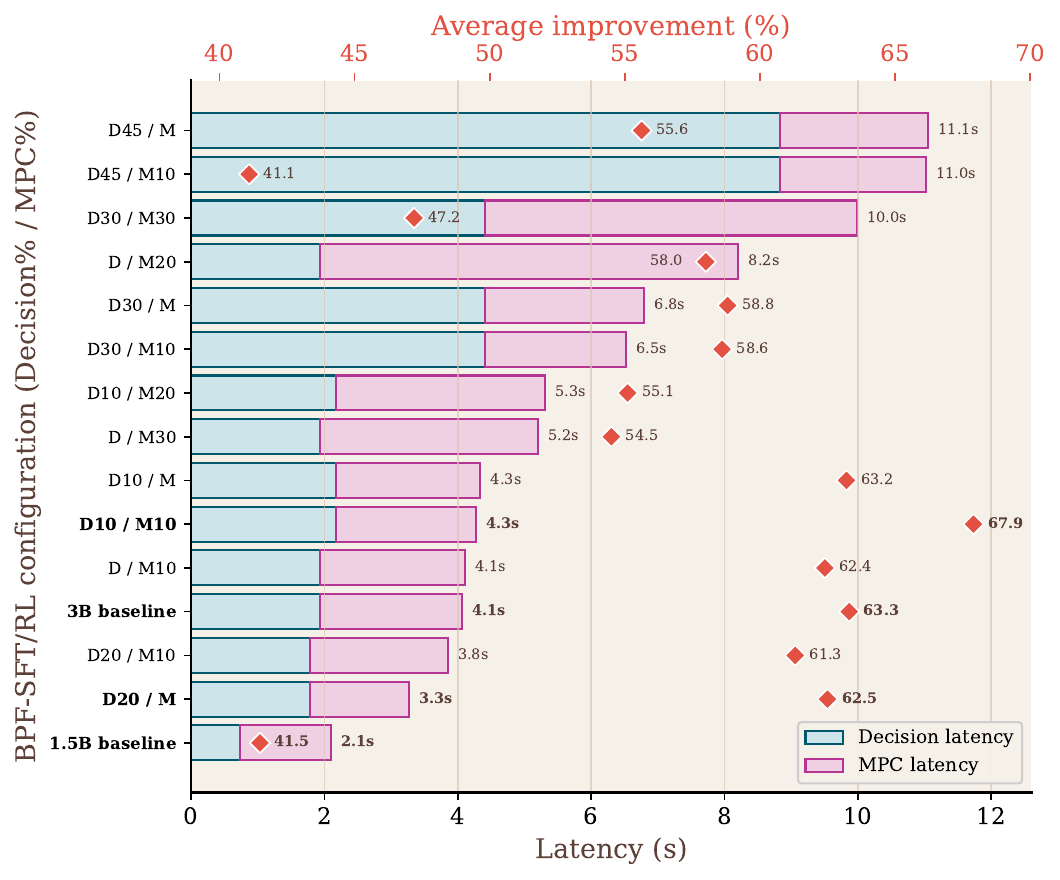} 
    \vspace{-2.0em}
    \caption{End-to-end latency of the full pipeline. The bottom x-axis reports measured latency, while the top x-axis reports corresponding task-performance from Fig.~\ref{fig:decision_mpc_pareto}. Pareto configurations are highlighted in \textbf{bold}.} 
    \label{fig:pipeline_latency} 
\end{wrapfigure}

While Table~\ref{tab:deployment} reports isolated-model throughput, the robot runs DecisionxR1 and MPCxR1 sequentially. We evaluate the complete cascade latency for the BPF-SFT/RL configurations from Sec.~\ref{subsec:dec_mpc_pruning_res}, measuring the number of tokens generated by each stage. Fig.~\ref{fig:pipeline_latency} shows that reducing parameters alone does not necessarily reduce end-to-end latency. This effect is not captured by isolated-model throughput and could be mitigated by training or prompting the modules to produce shorter outputs, for example, by asking MPCxR1 to generate only the parameters that need to be changed.
\begin{wraptable}{l}{0.5\textwidth}
    \centering
    \vspace{-1.0em}
    \caption{
    Embedded deployment metrics on the Jetson AGX Orin.
    }
    \label{tab:deployment}
    \resizebox{\linewidth}{!}{
    \begin{tabular}{c l c c}
    \toprule
    \textbf{Module} & \textbf{Configuration} & \textbf{Mem. (GB)} & \textbf{Dec. tok/s} \\
    \midrule
    \multirow{2}{*}{\rotatebox{90}{}} 
    & Baseline 3B   & 2.07 & 45 \\
    & Baseline 1.5B & 1.04 & 73 \\
    \midrule
    \multirow{4}{*}{\rotatebox{90}{Decision}} 
    & BPF-10\% & 1.92 & 49 \\
    & BPF-20\% & 1.77 & 51 \\
    & BPF-30\% & 1.62 & 56 \\
    & BPF-45\% & 1.51 & 58 \\
    \midrule
    \multirow{3}{*}{\rotatebox{90}{MPC}} 
    & BPF-10\% & 1.92 & 48 \\
    & BPF-20\% & 1.77 & 51 \\
    & BPF-30\% & 1.62 & 55 \\
    \bottomrule
    \end{tabular}
    }
\end{wraptable}
\luca{In this pipeline, pruning DecisionxR1 can change the text passed to MPCxR1, while pruning MPCxR1 can directly affect its output length. In both cases, longer outputs can reduce or completely hide the computational savings of a smaller model. A possible source of this effect is the separate adaptation of the two modules, since MPCxR1 is trained on outputs from the original, unpruned DecisionxR1. Notably, the D/M20 configuration, which prunes only MPCxR1 by $20\%$, doubles MPCxR1 latency from approximately $4.1$\,s to $8.2$\,s. This shows that upstream distribution shift from a pruned DecisionxR1 is not the sole cause of the increased verbosity, although separate module training may still contribute to it.}
The best latency-wise BPF-SFT/RL configuration prunes DecisionxR1 by $20\%$ and keeps MPCxR1 unpruned, reducing latency by $19.5$\% with only a $0.8$\% drop in task-performance.
Compared with the 1.5B baseline, this configuration is $1.57\times$ slower but achieves a $21.0$\% higher task-performance.

\section{Conclusions}

We introduced \emph{Before Parc Fermé} (BPF), a strategy that integrates pruning into the training process of embodied LLM controllers, with a particular focus on RL-time compression.
Across the considered robotic LLM pipeline, BPF-SFT/RL offers the best trade-off among the evaluated pruning schedules, demonstrating that the timing of compression matters for preserving closed-loop behavior.
Our deployment results show that structured pruning can reduce the memory footprint and decode throughput on the target robotic platform.
However, smaller modules are not necessarily faster end-to-end: pruning can change generated-token length and make the cascaded pipeline slower despite reducing parameter count.
This highlights that compression for modular robotic LLM systems must be evaluated at the full-pipeline level, jointly considering task performance, model size, generated output length, throughput, and latency.

\newpage
\section*{Limitations}
\label{sec:limitations}

This work presents \luca{five} main limitations.
First, the two LLM modules are adapted and pruned separately: MPCxR1 is trained on outputs from the original pipeline, not from pruned DecisionxR1 models.
This may reduce robustness to distribution shifts introduced by upstream pruning and may contribute to the increased verbosity and latency observed for some pruned cascades, motivating future work on joint or alternating training of compressed modules.
Second, we evaluate a limited set of pruning schedules, with fixed pruning intervals and fixed pruning increments; more frequent, smaller, or adaptive RL-time pruning policies remain unexplored.
Third, our structured pruning procedure is not quantization-aware, requiring padding to align pruned MLP dimensions with runtime constraints, which could be avoided by jointly optimizing pruning and quantization.
\luca{Fourth}, the RL objective does not penalize the length of generated tokens, although our deployment results show that pruning can increase verbosity and end-to-end latency. 
\luca{Finally, our evaluation is limited to the RobotxR1 pipeline with Qwen2.5 models in an autonomous-driving setting. Although the proposed BPF pruning schedules do not rely on RobotxR1-specific model architecture, whether the effectiveness of the approach holds across other embodied-LLM or VLA pipelines remains to be evaluated.}
\clearpage
\acknowledgments{This publication is part of the project PNRR-NGEU which has received funding from the MUR - DM 117/2023.}


\bibliography{bibliography}  
\newpage
\appendix
\section*{Appendix}

\section{RobotxR1 Pipeline and Evaluation Details}
\label{app:robotxr1_details}

This work uses the RobotxR1 pipeline~\cite{robotxr1} without modifying the controller, prompts, rewards, or evaluation protocol.
RobotxR1 builds on a 1:10 scaled racecar autonomy stack~\cite{forzaeth}, where a kinematic MPC tracks a minimum-curvature racing line in curvilinear coordinates.
The robot state contains the longitudinal and lateral displacement with respect to the racing line, heading error, steering angle, and longitudinal velocity.
The MPC exposes cost weights and constraints, such as velocity bounds and track-boundary inflation, which are the quantities updated by the LLM-generated parameters.

The pipeline consists of two RAG-enhanced LLM modules executed sequentially.
DecisionxR1 receives the user command $c$, the recent robot state history $h_t$, and auxiliary context $z_D$, including retrieved memories and formatting instructions:
\begin{equation}
    y_D = f_{\theta_D}(c, h_t, z_D),
    \label{eq:app_decision_llm}
\end{equation}
where $\theta_D$ denotes the DecisionxR1 weights.
The module determines whether the current vehicle behavior matches the requested behavior.
Conditioned on this output, MPCxR1 generates a textual update to the MPC parameters:
\begin{equation}
    y_M = f_{\theta_M}(c, h_t, y_D, z_M),
    \label{eq:app_mpc_llm}
\end{equation}
where $z_M$ includes the MPC formulation, retrieved memories, and output constraints.
The parsed parameters from $y_M$ are applied to the low-level MPC controller and evaluated in closed-loop simulation.

Following RobotxR1, control adaptability is measured as the percentage improvement over the default MPC controller across four behavior-specific task errors: centerline tracking error $E_c$, velocity-tracking error $E_v$, reversing error $E_r$, and smooth-driving error $E_s$.
The centerline error $E_c$ measures deviation from the centerline over a lap; $E_v$ measures deviation from the target velocity specified in the prompt; $E_r$ measures deviation from the reverse-driving target velocity; and $E_s$ measures deviation from zero acceleration, used as a proxy for smooth driving.
For each error $E$, the improvement is computed relative to the default MPC error, and the final score is the average over the four errors:
\begin{equation}
    \mathrm{AvgImprove}
    =
    \frac{100}{4}
    \sum_{E \in \{E_c,E_v,E_r,E_s\}}
    \frac{E^{\mathrm{MPC}} - E^{\mathrm{LLM}}}{E^{\mathrm{MPC}}}.
    \label{eq:app_avg_improvement}
\end{equation}
Here, $E^{\mathrm{MPC}}$ denotes the error obtained with the default MPC parameters, while $E^{\mathrm{LLM}}$ denotes the error after applying the LLM-generated MPC update.
The closed-loop evaluation uses the same randomized evaluation prompts, simulation setup, and Grand Tour race-track layout as RobotxR1.
The evaluation prompts are unseen during SFT and RL training and include perturbed natural-language variants for each behavior category.
\luca{For the results in Figs.~\ref{fig:decision_pareto}-\ref{fig:decision_mpc_pareto}, each configuration is evaluated over 1,576 randomized prompts, corresponding to approximately 25 evaluations for each combination of the eight robot states and eight human requests. All closed-loop control results are obtained in the ROS simulator on the Grand Tour track, while Sec.~\ref{subsec:deployment} reports deployment measurements on the Jetson AGX Orin mounted on the physical platform.}

Both modules are adapted through SFT followed by GRPO-based RL.
During SFT, the models are trained with the standard next-token prediction loss on task-specific instruction-response data.
During RL, DecisionxR1 uses behavior-adherence and formatting rewards, while MPCxR1 combines closed-loop driving-performance, formatting, and parameter-validity rewards.

\section{Experimental Setup}
\label{app:experimental_setup}

Both LLM modules, DecisionxR1 and MPCxR1, are instantiated from Qwen2.5~\cite{hui2024qwen2}.
The dense baselines are Qwen2.5-3B and Qwen2.5-1.5B models trained with the standard RobotxR1 SFT--GRPO pipeline; all pruning experiments start from the Qwen2.5-3B checkpoint.
SFT is run for 200 steps, with pruning, if applied, after 100 steps.
RL is run for 900 steps.
When pruning is applied during RL, we use a 400-step warm-up and split the target pruning ratio $\rho$ across three pruning events separated by $\Delta t_{\mathrm{RL}}=100$ steps, following Sec.~\ref{subsec:bpf_pruning}.
For BPF-SFT/RL, the target pruning ratio $\rho$ is split evenly across the two stages: $\rho/2$ is applied during SFT, and the remaining $\rho/2$ is applied iteratively during RL.
All schedules use the same total number of SFT and RL steps.

\luca{The 100-step SFT and 400-step RL warm-up intervals are kept fixed across pruning ratios and schedules. We selected these values through preliminary experiments to allow the model to first adapt to the supervised task format and the closed-loop RL objective, respectively, before pruning is introduced.}

\luca{Pruning does not otherwise modify the GRPO procedure, reward functions, or optimization objective. At each RL pruning event, the policy and reference models are pruned identically, preserving matched architectures and keeping the KL term well-defined throughout training.}

\section{Additional Results}\label{app:results}

\subsection{Additional DecisionxR1 Pruning Results}
\label{app:structured}

\begin{wrapfigure}{r}{0.5\textwidth}
    \centering
    \vspace{-1.0em}
    \includegraphics[width=\linewidth]{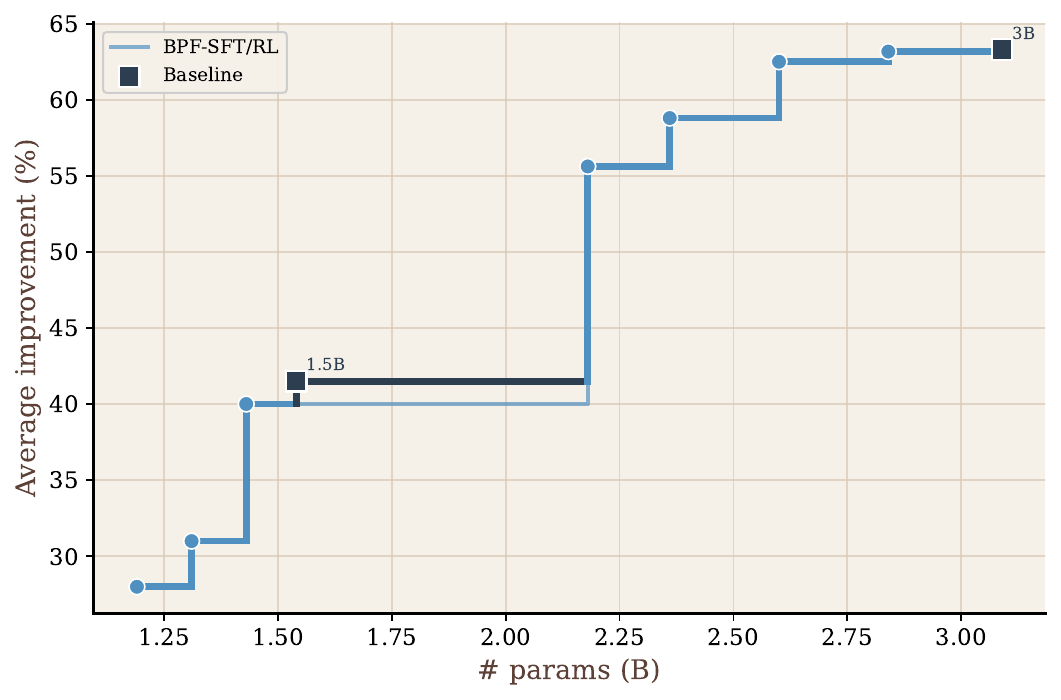}
    \vspace{-1.0em}
    \caption{
    DecisionxR1 pruning with fixed MPCxR1, using both 3B and 1.5B as starting checkpoints for BPF-SFT/RL.
    }
    \label{fig:app_decision_pareto_15b}
\end{wrapfigure}
Fig.~\ref{fig:app_decision_pareto_15b} extends the DecisionxR1 pruning analysis of Fig.~\ref{fig:decision_pareto} by also applying BPF-SFT/RL to the 1.5B checkpoint.
The results show that the smaller dense model has less pruning headroom.
Starting from 1.5B, a mild $10\%$ pruning ratio only slightly reduces average control adaptability, from $41.5\%$ to $40.01\%$.
However, at $20\%$ pruning the score drops to $31\%$, showing that the performance degradation appears much earlier than when pruning the 3B model.

\begin{wrapfigure}{r}{0.5\textwidth}
    \centering
    \vspace{-4.0em}
    \includegraphics[width=\linewidth]{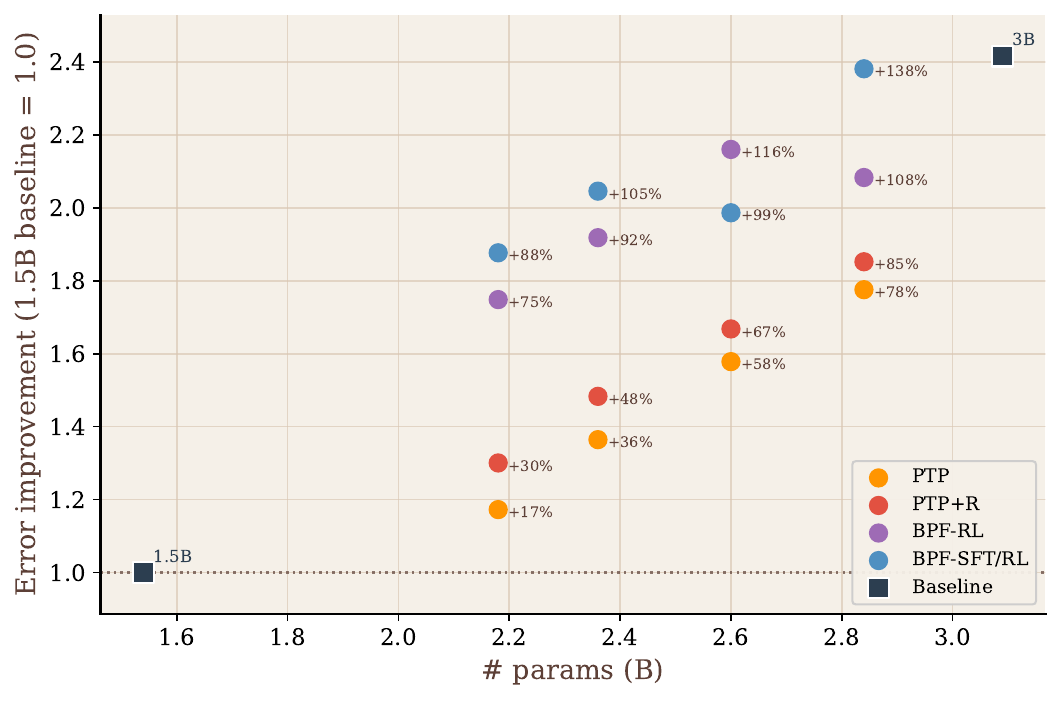}
    \caption{
    Average task-error improvement of pruned DecisionxR1 models initialized from 3B, measured relative to the 1.5B baseline.}
    \vspace{-1.0em}
    \label{fig:app_decision_error_improv}
\end{wrapfigure}

Fig.~\ref{fig:app_decision_error_improv} reports the same 3B-initialized pruning configurations in terms of average task-error improvement over the 1.5B baseline.
Only models initialized from 3B are included.
The plot confirms that BPF-based pruning remains consistently above the smaller dense baseline across pruning ratios.
At the $10\%$ pruning ratio, BPF-RL slightly outperforms BPF-SFT/RL on this error-improvement metric.
However, this local advantage does not translate into a better average control-adaptability score in Fig.~\ref{fig:decision_pareto}, where BPF-SFT/RL provides the strongest overall trade-off.

\subsection{Unstructured Pruning on DecisionxR1}\label{app:unstructured}
In the main paper, we focus on structured pruning because it yields smaller dense architectures and is therefore better aligned with standard inference runtimes. For completeness, we also evaluate unstructured pruning on DecisionxR1. This analysis is limited to DecisionxR1 accuracy and does not include end-to-end closed-loop evaluation with MPCxR1. Therefore, these results should be interpreted as a module-level robustness check rather than as evidence of full-pipeline control performance.

\begin{figure}
    \centering
    \begin{subfigure}[t]{0.48\linewidth}
        \centering
        \includegraphics[width=\linewidth]{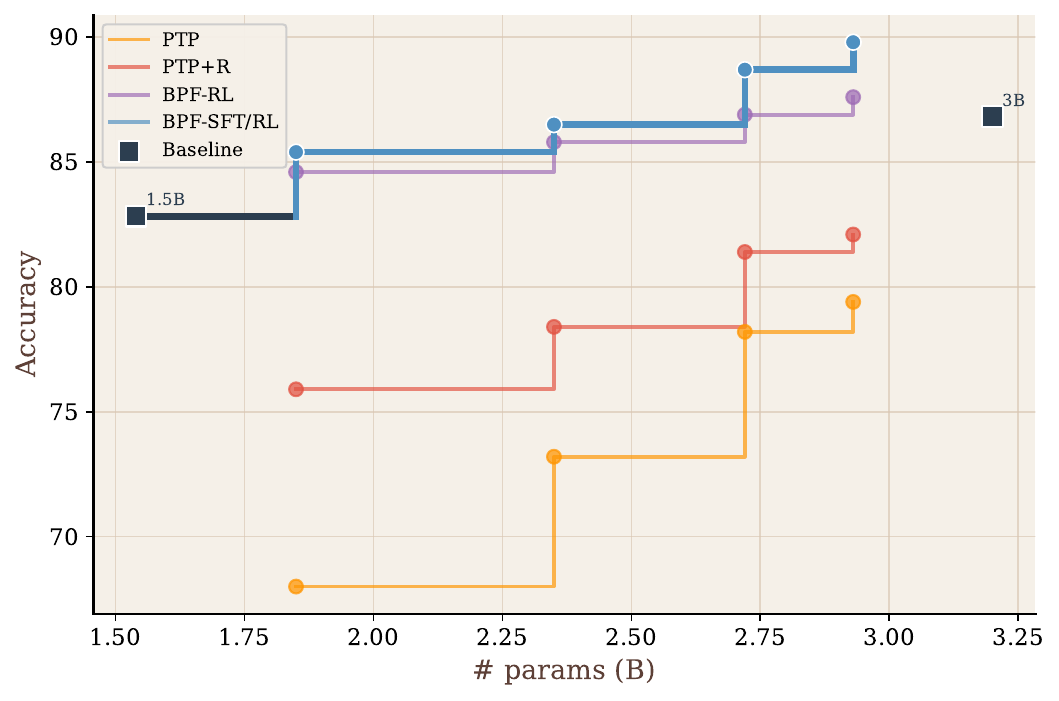}
        \caption{DecisionxR1 accuracy under unstructured pruning. All pruned models start from 3B, while 1.5B is shown as the smaller dense baseline.}
        \label{fig:unstructured_decision_accuracy_pareto}
    \end{subfigure}
    \hfill
    \begin{subfigure}[t]{0.48\linewidth}
        \centering
        \includegraphics[width=\linewidth]{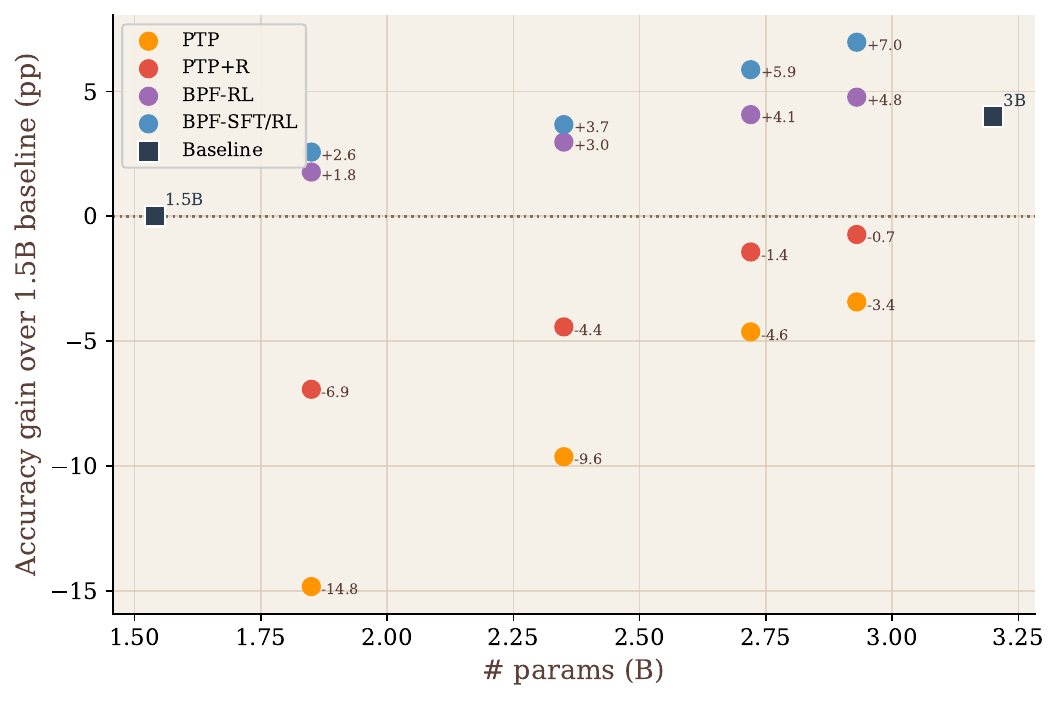}
        \caption{Relative accuracy improvement of unstructured pruned DecisionxR1 models over the dense 1.5B baseline. 
        }
        \label{fig:unstructured_decision_gain_15b}
    \end{subfigure}
    \caption{DecisionxR1 results under unstructured pruning.}
    \label{fig:unstructured_pruning}
\end{figure}

Fig.~\ref{fig:unstructured_decision_accuracy_pareto} shows that unstructured pruning preserves DecisionxR1 accuracy over a wide range of parameter budgets.
In particular, the pruned 3B-derived BPF models remain above the dense 1.5B baseline for moderate and high pruning ratios.
BPF-SFT/RL reaches $89.80\%$ accuracy with $2.93$B parameters, slightly outperforming the unpruned 3B DecisionxR1 baseline on this module-level metric.
Even at stronger compression, with $1.85$B parameters, BPF-SFT/RL reaches $85.40\%$ accuracy, remaining above the $82.83\%$ accuracy of the dense 1.5B baseline.

Fig.~\ref{fig:unstructured_decision_gain_15b} reports the same results as relative accuracy improvement over the dense 1.5B baseline.
The improvement remains positive across the reported unstructured BPF configurations, indicating that the larger 3B backbone retains useful redundancy even after a substantial fraction of weights is removed.
Interestingly, the post-training pruning variants, PTP and PTP+R, fall below the dense 1.5B baseline, showing that post-training pruning can degrade DecisionxR1 accuracy below the smaller dense baseline when compression is not integrated into adaptation.
However, unlike structured pruning, unstructured sparsity does not directly produce a smaller dense architecture and may require sparse kernels or hardware support to translate parameter sparsity into practical inference speedups.
For this reason, we use structured pruning for the main end-to-end and deployment experiments, while reporting unstructured pruning here as a module-level robustness check.

\subsection{Pruning-Criterion Sensitivity}
\label{app:wanda}

\begin{wrapfigure}{r}{0.5\textwidth}
    \centering
    \vspace{-4.0em}
    \includegraphics[width=\linewidth]{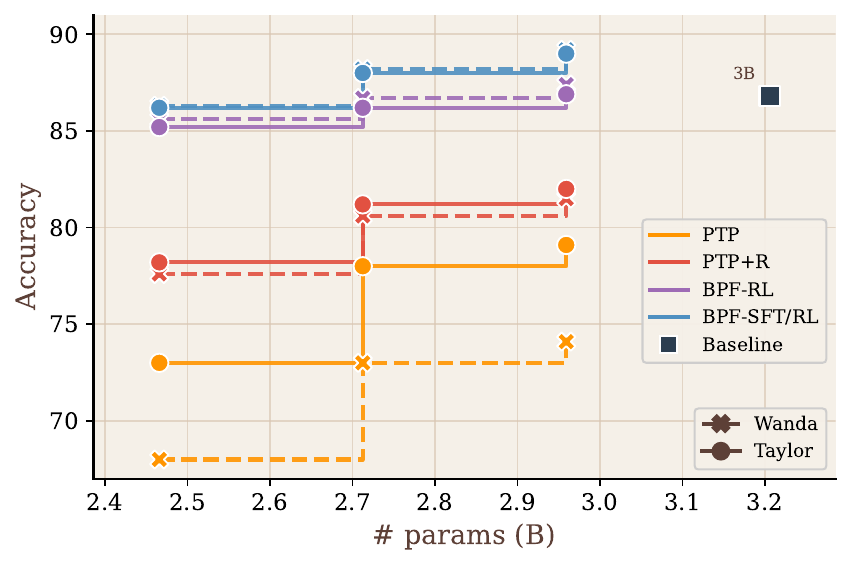}
    \caption{DecisionxR1 accuracy under matched pruning schedules using first-order Taylor scoring and Wanda.}
    \vspace{-2.0em}
    \label{fig:taylor_wanda}
\end{wrapfigure}

The main experiments of this manuscript evaluate the proposed pruning schedules using the first-order Taylor criterion described in Sec.~\ref{subsec:taylor_pruning}. To assess whether the observed scheduling trends depend on this criterion, we additionally evaluate Wanda~\cite{wanda} on DecisionxR1. Since Wanda applies unstructured pruning, we report DecisionxR1 accuracy, as in Sec.~\ref{app:unstructured}, rather than end-to-end deployment metrics.

Fig.~\ref{fig:taylor_wanda} compares Taylor and Wanda under the same pruning schedules. The same overall trend is observed with both criteria: PTP performs worst, RL recovery partially recovers the lost accuracy, and the RL-coupled schedules generally provide the strongest results.

\newpage
\subsection{Iterative SFT Pruning}
\label{app:iter_sft}

\begin{wrapfigure}{r}{0.5\textwidth}
    \centering
    \vspace{-4.0em}
    \includegraphics[width=\linewidth]{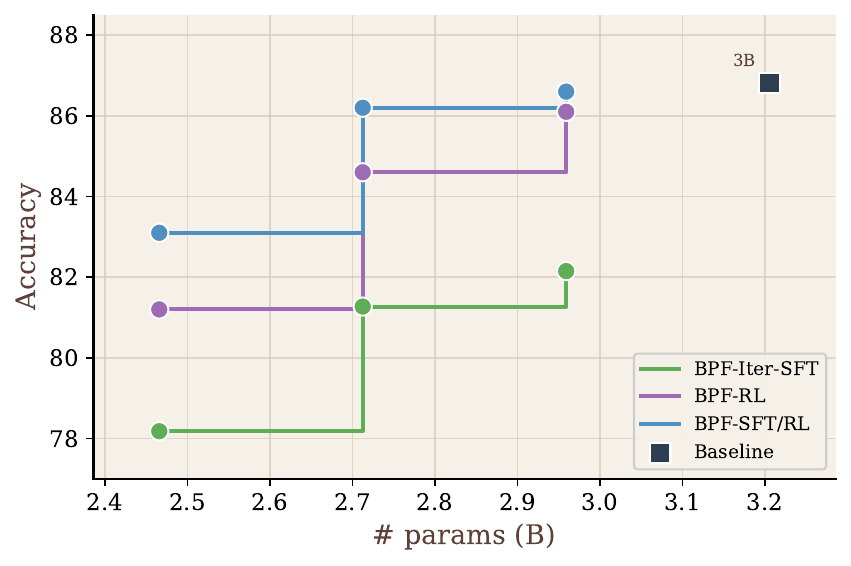}
    \caption{DecisionxR1 accuracy for iterative SFT pruning compared with BPF-RL and BPF-SFT/RL.
    }
    \vspace{-1.0em}
    \label{fig:iter_sft}
\end{wrapfigure}

To separate the effect of iterative pruning from its placement during RL, we introduce an additional iterative SFT-only baseline. After the same 100-step SFT warm-up used in the other schedules, the target pruning ratio is split across three pruning events distributed over the remaining SFT steps, with supervised updates between consecutive pruning events. The subsequent RL stage proceeds without further pruning. Thus, this baseline retains incremental pruning and intermediate adaptation, but all pruning events occur during SFT rather than RL.

Fig.~\ref{fig:iter_sft} compares this schedule with BPF-RL and BPF-SFT/RL. Iterative SFT pruning consistently underperforms BPF-SFT/RL across the considered pruning ratios. Therefore, the improvement of BPF-SFT/RL cannot be explained only by splitting compression across multiple pruning events. Instead, the results support coupling pruning with RL optimization, where adaptation between pruning events is guided by task-level feedback.

\subsection{Multi-Seed Robustness}
\label{app:seeds}

\begin{wraptable}{l}{0.5\textwidth}
    \centering
    \caption{Results over four training seeds.}
    \label{tab:seed_results}
    \small
    \setlength{\tabcolsep}{4pt}
    \renewcommand{\arraystretch}{0.9}
    \begin{tabular}{c l c}
    \toprule
    \textbf{Metric} & \textbf{Configuration} & \textbf{Result (\%)} \\
    \midrule
    E2E impr.
    & D10/M10, BPF-SFT/RL & $66.1 \pm 5.1$ \\
    \midrule
    \multirow{5}{*}{\rotatebox[origin=c]{90}{Dec. acc.}}
    & Dense 3B            & $86.4 \pm 1.3$ \\
    & D45/M0, PTP         & $57.2 \pm 3.4$ \\
    & D45/M0, PTP+R       & $69.1 \pm 3.9$ \\
    & D45/M0, BPF-RL      & $75.6 \pm 4.7$ \\
    & D45/M0, BPF-SFT/RL  & $\mathbf{79.5 \pm 4.1}$ \\
    \bottomrule
    \end{tabular}
\end{wraptable}
To assess training variability, we repeat four independent training runs for the key D10/M10 end-to-end configuration and for the D45 DecisionxR1 comparison across pruning schedules. Table~\ref{tab:seed_results} reports mean and standard deviation across training seeds.

For D45, the original ordering across pruning schedules is preserved: BPF-SFT/RL achieves the highest DecisionxR1 accuracy, followed by BPF-RL, PTP+R, and PTP. The D10/M10 experiment further shows that the configuration remains stable across independent training runs, reaching $66.1\pm5.1\%$ average improvement.

\subsection{Qualitative Trajectories}
\label{app:qualitative_decision}

\begin{wrapfigure}{r}{0.42\textwidth}
    \centering
    \vspace{-4.0em}
    \includegraphics[width=\linewidth]{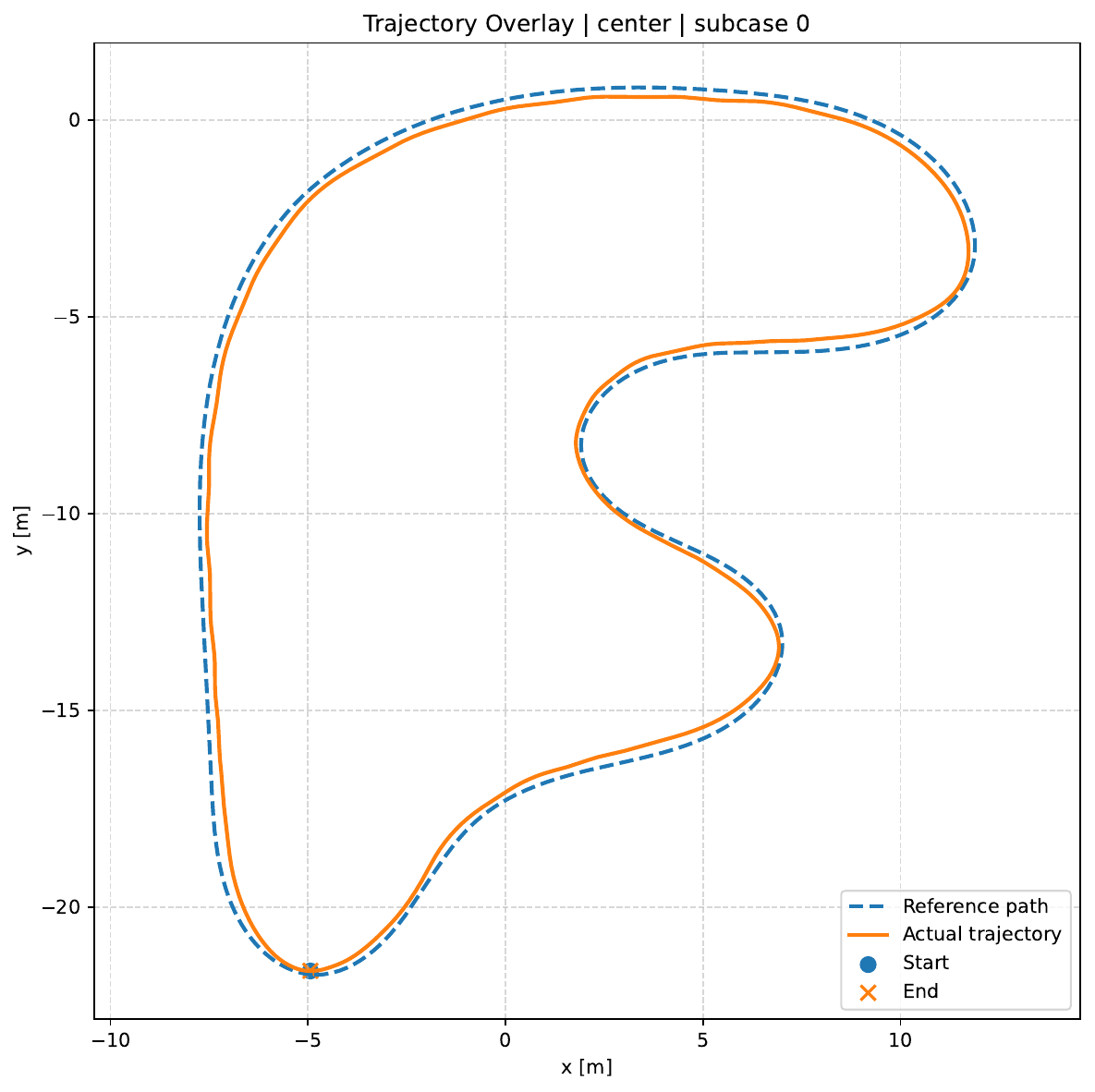}
    \caption{
    Original unpruned DecisionxR1-MPCxR1 trajectory for the centerline-following prompt.
    }
    \label{fig:qualitative_decision_original}
    \vspace{-2.0em}
\end{wrapfigure}

\begin{figure*}[t]
    \centering
    \includegraphics[width=0.24\linewidth]{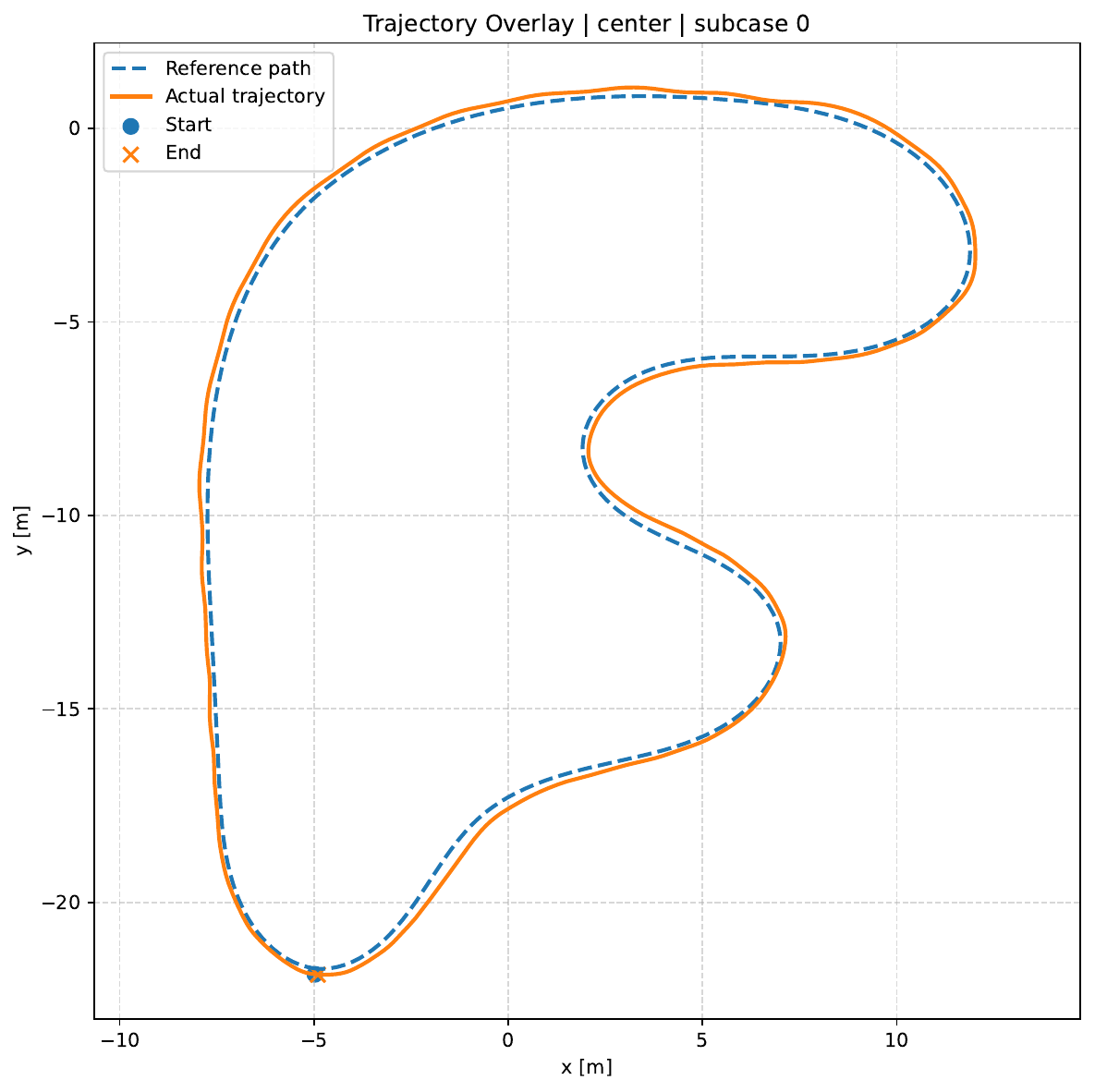}
    \includegraphics[width=0.24\linewidth]{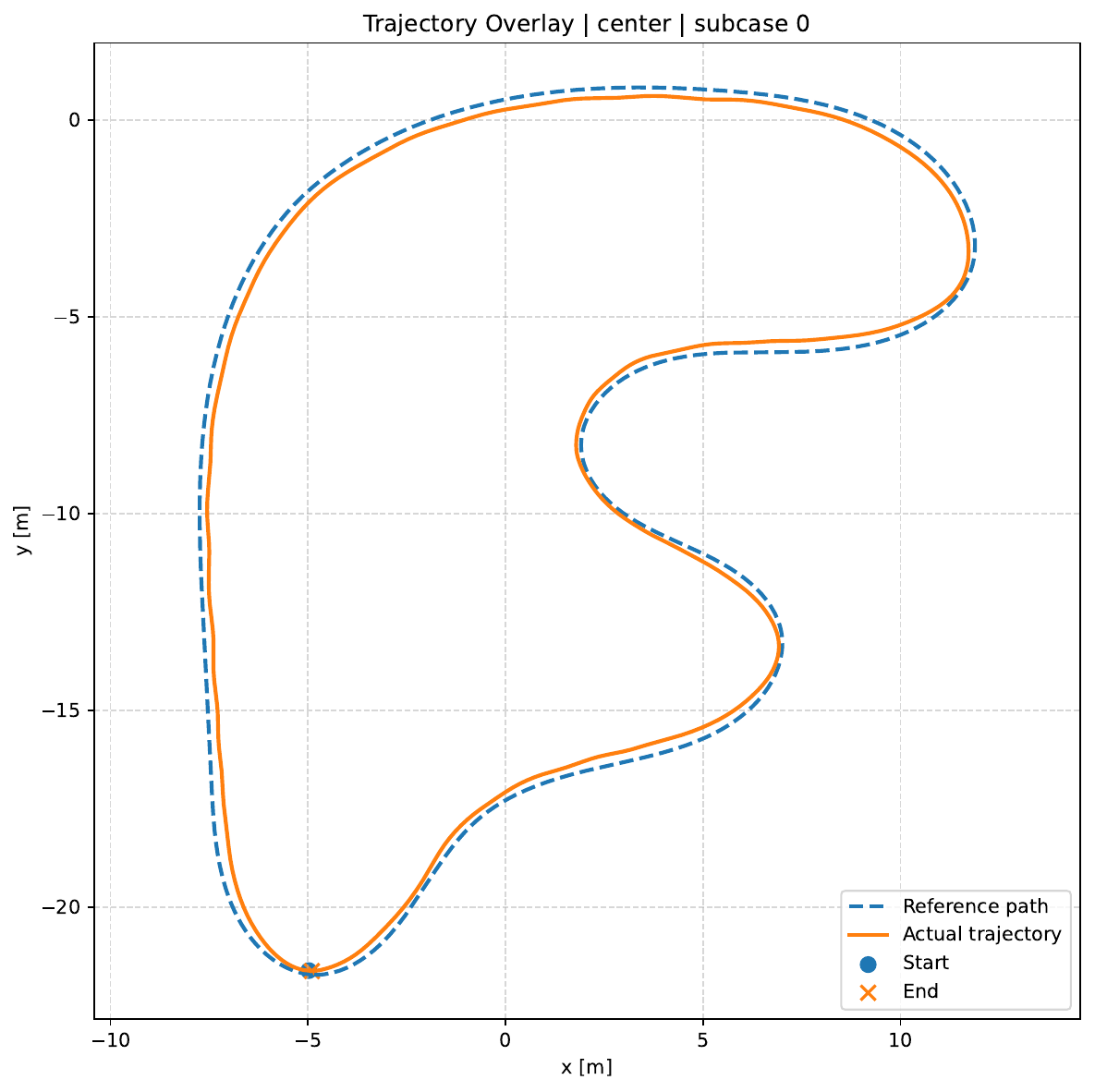}
    \includegraphics[width=0.24\linewidth]{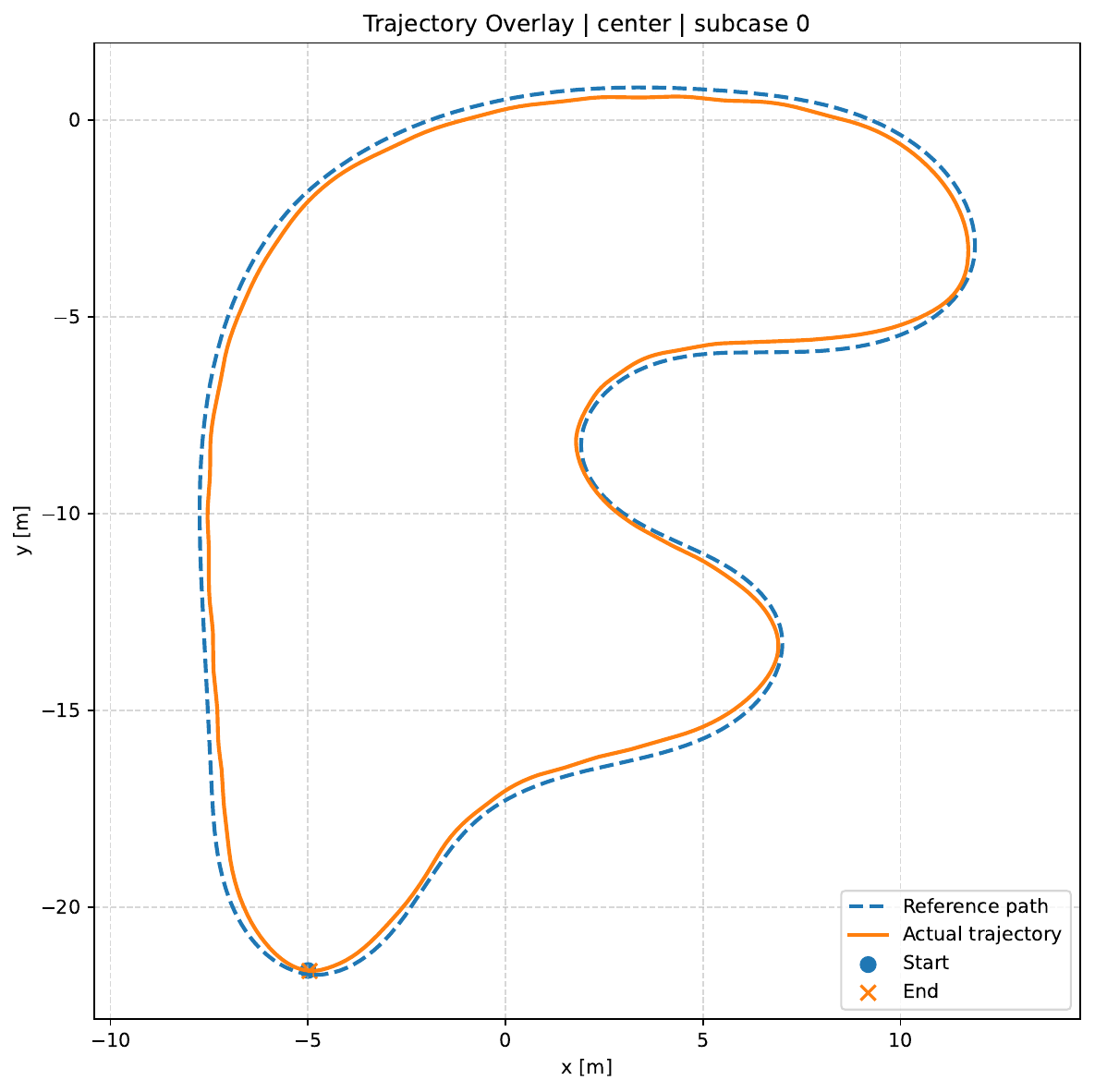}
    \includegraphics[width=0.24\linewidth]{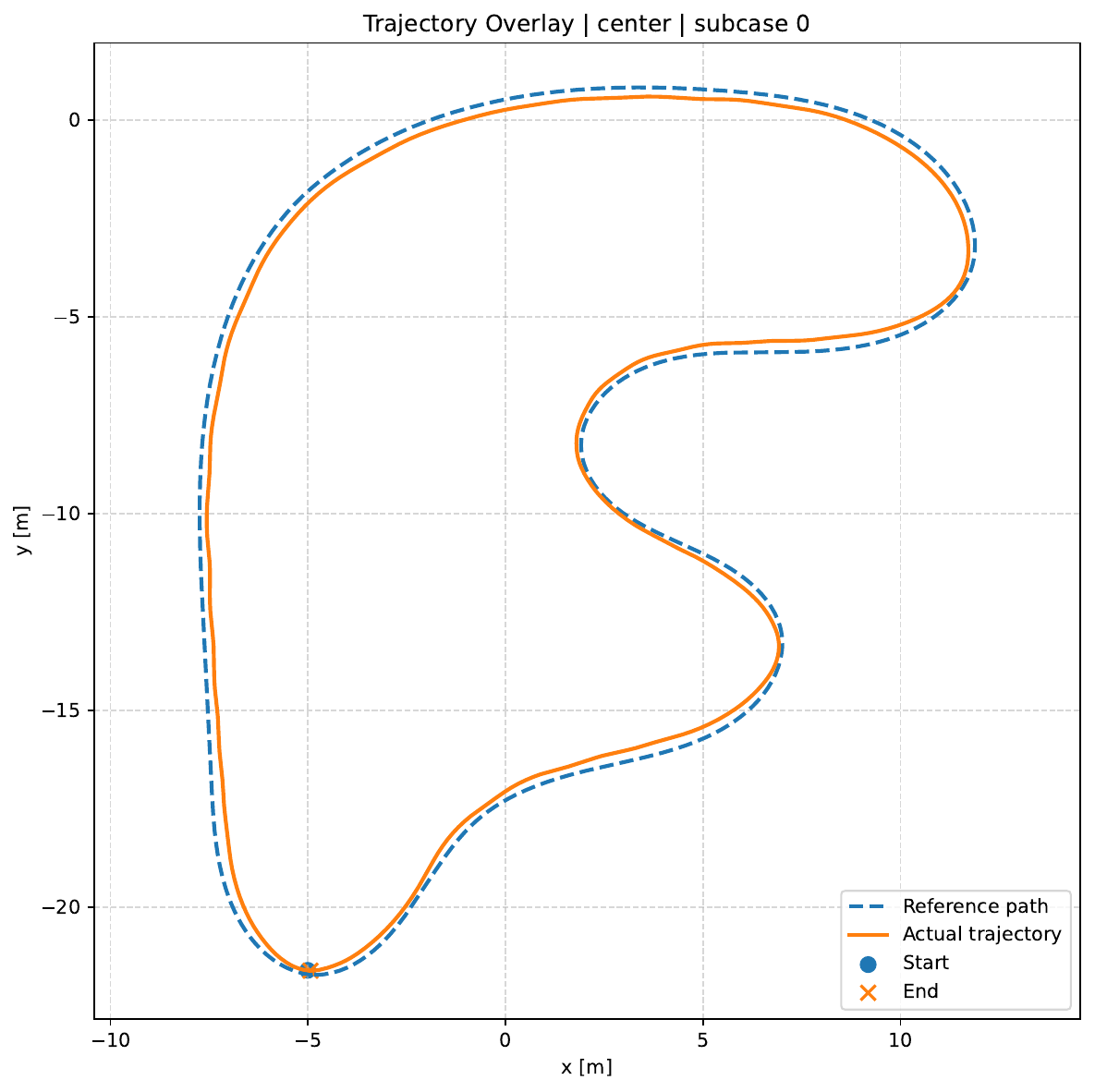}
    \caption{
    Qualitative ROS trajectories for DecisionxR1 pruned at $10\%$, $20\%$, $30\%$, and $45\%$ with BPF-SFT/RL, keeping MPCxR1 unpruned.
    }
    \label{fig:qualitative_decision_pruning}
\end{figure*}

To complement the quantitative results, Fig.~\ref{fig:qualitative_decision_original} and Fig.~\ref{fig:qualitative_decision_pruning} show qualitative closed-loop trajectories in the ROS simulator for the prompt \emph{``Stay directly on the middle of the track''}. 
We compare the original unpruned DecisionxR1-MPCxR1 pipeline (Fig.~\ref{fig:qualitative_decision_original}) against four configurations where MPCxR1 is kept unpruned and DecisionxR1 is pruned with BPF-SFT/RL to $10\%$, $20\%$, $30\%$, and $45\%$, respectively (Fig.~\ref{fig:qualitative_decision_pruning}).
In each case, the figure overlays the reference path and the trajectory followed by the vehicle.

The results show that the trajectories produced by the pruned DecisionxR1 models remain visually close to the unpruned trajectory in Fig.~\ref{fig:qualitative_decision_original}.
Even at $45\%$ pruning, the vehicle still follows the centerline behavior with a trajectory comparable to the unpruned baseline.
These qualitative results are consistent with the end-to-end control-adaptability scores reported in Sec.~\ref{subsec:decision_pruning_fixed_mpc}, indicating that the proposed pruning strategy preserves the relevant driving behavior despite substantially reducing the DecisionxR1 model size.

\subsection{Deployment on Multiple Reference Devices}
\label{app:deployment}

Table~\ref{tab:deployment_all} extends the deployment results of Sec.~\ref{sec:Results} by reporting both prefill and decode throughput on two devices: the Jetson AGX Orin onboard the robotic platform and an RTX A5000 workstation GPU.
All models use the same deployment pipeline described in Sec.~\ref{sec:Results}: BPF-SFT/RL-pruned checkpoints are quantized to \texttt{Q5\_K\_M}, exported to \texttt{GGUF}, and executed with \texttt{llama.cpp}.
As in the main text, the dense baselines are re-measured on our deployment stack.

\begin{table}[ht]
\centering
\caption{
Deployment metrics for dense and BPF-SFT/RL-pruned models on an RTX A5000 workstation GPU and on the Jetson AGX Orin onboard the robotic platform.
}
\label{tab:deployment_all}
\resizebox{\linewidth}{!}{
\begin{tabular}{c l c c c c c}
\toprule
\multirow{2}{*}{\textbf{Module}} &
\multirow{2}{*}{\textbf{Configuration}} &
\multirow{2}{*}{\textbf{Memory (GB)}} &
\multicolumn{2}{c}{\textbf{RTX A5000}} &
\multicolumn{2}{c}{\textbf{Jetson AGX Orin}} \\
\cmidrule(lr){4-5}
\cmidrule(lr){6-7}
&
&
&
\textbf{Prefill tok/s} &
\textbf{Decode tok/s} &
\textbf{Prefill tok/s} &
\textbf{Decode tok/s} \\
\midrule
\multirow{2}{*}{\rotatebox{90}{}} 
& Baseline 3B   & 2.07 & 7303  & 188 & 1744 & 45 \\
& Baseline 1.5B & 1.04 & 10865 & 270 & 3088 & 73 \\
\midrule
\multirow{4}{*}{\rotatebox{90}{Decision}} 
& BPF-10\% & 1.92 & 7514 & 194 & 1863 & 49 \\
& BPF-20\% & 1.77 & 8233 & 207 & 1894 & 51 \\
& BPF-30\% & 1.62 & 7903 & 209 & 2073 & 56 \\
& BPF-45\% & 1.51 & 9525 & 225 & 2184 & 58 \\
\midrule
\multirow{3}{*}{\rotatebox{90}{MPC}} 
& BPF-10\% & 1.92 & 7890 & 197 & 1784 & 48 \\
& BPF-20\% & 1.77 & 7092 & 200 & 1985 & 51 \\
& BPF-30\% & 1.62 & 8262 & 210 & 2124 & 55 \\
\bottomrule
\end{tabular}
}
\end{table}

\end{document}